# Personalized Federated Learning: A Combinational Approach

Sone Kyaw Pye
School of Computer Science and Engineering
Nanyang Technological University
Singapore
pyesone001@e.ntu.edu.sg

Han Yu
School of Computer Science and Engineering
Nanyang Technological University
Singapore
han.yu@ntu.edu.sg

## ABSTRACT

Federated learning (FL) is a distributed machine learning approach involving multiple clients collaboratively training a shared model. Such a system has the advantage of more training data from multiple clients, but data can be non-identically and independently distributed (non-i.i.d.). Privacy and integrity preserving features such as differential privacy (DP) and robust aggregation (RA) are commonly used in FL. In this work, we show that on common deep learning tasks, the performance of FL models differs amongst clients and situations, and FL models can sometimes perform worse than local models due to non-i.i.d. data. Secondly, we show that incorporating DP and RA degrades performance further. Then, we conduct an ablation study on the performance impact of different combinations of common personalization approaches for FL, such as finetuning, mixture-of-experts ensemble, multi-task learning, and knowledge distillation. It is observed that certain combinations of personalization approaches are more impactful in certain scenarios while others always improve performance, and combination approaches are better than individual ones. Most clients obtained better performance with combined personalized FL and recover from performance degradation caused by non-i.i.d. data, DP, and RA.

## CCS CONCEPTS

• Computing methodologies ~ Artificial intelligence ~ Distributed artificial intelligence

## 1 Introduction

Federated Learning (FL) is a distributed machine learning (ML) approach that involves multiple users, referred to as clients, collaboratively training a global model without transferring data from local storage to a central server [1]. Ideally, an FL model performs better than individual models trained only on each client's data due to more training data. FL can be further classified into two scenarios: cross-device and cross-silo FL, with the difference between them being the number of clients, with the latter having significantly fewer clients but with more data per client [2]. FL's distributed data paradigm contrasts traditional ML, which requires data to be stored in a single location which brings about concerns in terms of communication costs and privacy. There can be significant communication overheads incurred in transferring data from devices to a central location, and such data movement might contravene privacy laws such as the European General Data Protection Regulation [3] and Health Insurance Portability and Accountability Act [4]. FL addresses these concerns as there is no need for centralized data storage, causing it to garner increasing interest recently, resulting in various emerging FL applications across multiple fields.

Two challenges in FL are the non-identically and independently distributed (non-i.i.d.) nature of data and model heterogeneity caused by dissimilar clients needing specific models for their unique context. These challenges cause the FL model to perform poorly for certain clients as it fails to adapt to unique distributions of data of individual clients, resulting in local models for clients performing better than an FL model [5], reducing incentive for those clients to participate in FL. FL also faces privacy concerns as FL models have been shown to leak training data [6]. To counteract this, differential privacy (DP) has been proposed. FL is also vulnerable to adversarial attacks, and robust aggregation (RA) has been suggested as a defense against some attacks [7]. However, incorporating DP and RA degrades performance, which disincentivizes clients to participate in FL since there might not be a significant or even a positive performance gain with these additional measures in place.

One way to overcome performance degradation is the personalization of FL models. There are several personalization approaches such as finetuning (FT), mixture of experts (MoE), multi-task learning (MTL), and knowledge distillation (KD), each of which has been shown to yield positive performance gains for FL models [7]. However, to our knowledge, no work has been done to combine personalization approaches, which could provide further performance gains.

This paper addresses this gap and evaluates various combinations of personalization approaches in scenarios of plain FL, differentially private FL (DP-FL), and robust aggregation FL (RA-FL). We observe that existing personalization approaches affect different aspects of the FL process, such as MoE not affecting the FL model, while FT further trains the FL model after federated training, and KD and MTL modifies FT. Therefore, we try out all possible combinations of these approaches. Our main contributions in this paper are as follows:

- We demonstrate that for certain clients, FL does not provide enough performance incentive to be a part of the federation of clients and incorporating DP and RA can further reduce that incentive due to performance degradation.
- We propose combinations of personalization approaches comprising common personalization approaches.

- We empirically show that these combinations yield better performance gains than standalone personalization approaches and compensate for performance degradation.
- We observe that certain combinations are more impactful in certain scenarios and tasks while others improve performance across the board, and a combination of approaches always tends to be better than individual ones.

The rest of the paper is organized as follows. Section 2 presents background on FL, DP, RA, and personalization approaches. Section 3 presents our experimental setup, Section 4 showcases results and analysis and conclusion is presented in Section 5.

## 2 Background

### 2.1 Federated Learning

A typical FL training process would encompass:
1. **Selection of training participants**: In each FL training round $t = 1 \ldots T$, server randomly samples $m$ clients. This selection only occurs in cross-device FL, as cross-silo FL involves all clients due to the small number of total clients.
2. **Distribution of Initial Global Model**: Selected clients from Step 1 download the latest model $G^{t-1}$ from server.
3. **Local Training**: Each client trains $G^{t-1}$ for K epochs using local data and computes an update $P_i^t$ to $G^{t-1}$.
4. **Aggregation**: Server collects updates and averages them into updated global model $G^t$ using Federated Averaging (FedAvg) with aggregation learning rate η:

$$G^t = G^{t-1} + \frac{\eta}{m}\sum_{i=1}^{m}(P_i^t - G^{t-1}) \quad (1)$$

This process can go on as long as new data is available for training from clients and there are clients eligible for training.

### 2.2 Differential Privacy & Robust Aggregation

DP limits information learnable about clients from model updates or FL models [8]. However, DP degrades performance of the FL model. In formal terms, differential privacy (DP) provides an $(\epsilon, \delta)$ privacy guarantee when the federated mechanism M and two sets of users Q, Q' that differ by one participant produce models in any set G with probabilities that satisfy:

$$\Pr[M(Q) \in G] \leq e^\epsilon \Pr[M(Q') \in G] + \delta \quad (2)$$

To incorporate DP in FL, it involves clipping each client's update and adding Gaussian noise $N(0, \sigma)$. Referencing Equation 1, aggregation is modified as follows:

$$G^t = G^{t-1} + \frac{\eta}{m}\sum_{i=1}^{m}(Clip(P_i^t - G^{t-1}, S)) + N(0, \sigma) \quad (3)$$

where $S$ is clipping bound, and $N(0, \sigma)$ is noise added. These values are dependent on the number of clients. The lesser the number of clients, the larger the magnitude of clipping and noise added to preserve privacy [8]. This makes it incompatible with cross-silo FL, with its small number of clients.

RA is a suggested defense against poisoning attacks of malicious clients. RA replaces FedAvg, and instead of averaging updates like FedAvg, the geometric median is used [9]. Typically, poisoning attacks would involve scaling model weights or using poisoned data to train the FL model before sending the poisoned updates for aggregation. RA reduces the impact that statistical outliers have on model weights as only median weight, which outliers do not contribute to, is used. RA is represented as such:

$$G^t = G^{t-1} + \eta(\tilde{P}^t - G^{t-1}) \quad (4)$$

where $\tilde{P}^t$ is the element-wise median for the updates acquired by the server performing the aggregation in round $t$ of FL training. RA has also been shown to degrade performance of models [7].

### 2.3 Personalization of FL Models

Numerous personalization approaches have been proposed, and most can be categorized into the following archetypes:

**Finetuning (FT)**: FL model after federated training is further trained on client's local data. The intuition is akin to transfer learning, where knowledge acquired from a global pool of data is leveraged to learn better local features instead of learning from scratch on a limited local pool of data [10]. A variant of FT, called **freeze-base FT**, involves freezing some model layers, such as base layers, and leaving only top layers unfrozen [7].

**Multi-task learning (MTL)**: A MTL problem involves solving multiple related tasks together using commonalities across tasks [11]. In FL, training the FL model and personalizing it can be treated as related tasks [12]. In [7], FL training process is treated as task X and personalization for a client as task Y to formulate an MTL problem. The aim is to use $G^T$, which is optimized for X, and optimize it for Y, producing personalized model A.

This optimization can be viewed as an extension of FT with a different loss function. To address possible catastrophic forgetting [13] of X while optimizing for Y, elastic weight controls [14] are used [7] to reduce rate of learning on critical layers/weights for X. As such, cross-entropy loss is augmented:

$$l(A, x) = L_{cross}(A, x) + \sum_i \left(\frac{\lambda}{2} F_i (A_i - G_i^T)^2\right) \quad (5)$$

where $\lambda$ is the importance of task X vs. Y, F is the Fisher information matrix, and $i$ is the label of each parameter.

**Knowledge distillation (KD)**: KD involves extracting learned features of a teacher model to teach a student model [15]. In FL, treating FL model ($G^T$) as the teacher and personalized model (A) as the student and using loss function from knowledge distillation literature, KD can be viewed as an extension to FT. Like MTL, the cross-entropy loss function is augmented as such:

$$l(A, x) = \alpha K^2 L_{cross}(A, x) \quad (6)$$
$$+ (1 - \alpha) KL(\sigma\left(\frac{G^T(x)}{K}\right), \sigma\left(\frac{A(x)}{K}\right))$$

where KL is Kullback-Leibler divergence loss, $\sigma$ is softmax, $\alpha$ is the weight parameter, and K is the temperature constant.

**Mixture of Experts (MoE)**: MoE treats personalization as an ensemble learning task, where a local model trained only on the client's data is used together with the FL model [16]. The local model and FL model are put into a weighted average ensemble, which can be expressed as such:

$$y = \alpha(G^T(x)) + (1 - \alpha)(DE_i(x))$$

where $x$ is the testing data, $G^T$ is the FL model, $DE_i$ is the domain expert, $\alpha$ is the weight, and $y$ is the prediction/output.

**Meta-learning**: A meta-learner trains a model on similar tasks with the aim of adapting to a new but similar task quickly despite limited data for the new task [17]. For FL, meta-learning considers personalization for clients as similar tasks [18].

Although these approaches have been studied individually, no study has been conducted to explore the efficacy of combining personalization approaches. As certain archetypes such as meta-learning are incompatible with others, they will not be included in our study. The compatible approaches (FT, MTL, KD, MoE) have not been studied on both cross-silo and cross-device FL scenarios, and a comparison across scenarios, tasks, and combinations of personalization approaches has not been done before.

## 3   Methodology

This study explores how different combinations of personalization approaches impacts performance of FL across various tasks/scenarios. Tasks and datasets used are in Table 1.

Table 1: FL Tasks and Datasets

| Scenario | Task | Dataset | Clients |
|---|---|---|---|
| Cross-Silo | Image Classification | Office | 3 |
| | | DomainNet | 5 |
| | Text Classification | Cross-Sector | 3 |
| | | Cross-Product | 5 |
| Cross-Device | Image Classification | CIFAR-10 | 100 |
| | Next Word Prediction | Reddit | 80,000 |

### 3.1   Datasets

As datasets made explicitly for FL are still rare, it is common to retrofit ML datasets into FL ones by dividing the datasets into subsets, representing clients. Domain adaptation datasets, which already have the dataset divided into domains, can be used, with each domain representing a client. The subsections that follow will elaborate on the details of the six datasets we used.

**Office Dataset:** This dataset contains three domains/clients: Amazon, Webcam, and DSLR, representing cross-silo FL scenario for image classification. Each client contains images from Amazon or images taken using a webcam or DSLR camera [19]. The unequal number and different origin of images for each client simulate the non-iid nature of FL data.

**DomainNet Dataset**: This dataset contains five domains/clients: Infograph, Painting, Quickdraw, Real and Sketch [20], with each client having different forms of visual representations for the same classes of objects. This dataset is used for the cross-silo FL scenario for the task of image classification. Only a subset of the entire DomainNet dataset was used for this project in the interest of time and available computation resources. The subset has seventeen randomly chosen classes but retains the five domains as per the full dataset.

**Cross-Product Dataset:** This dataset comprises five smaller consumer review datasets of different product categories, each acting as a client: Amazon-branded products, Alexa-branded products, food, phones, and headphones [21-25]. Reviews are rated 1 to 5. The datasets were obtained from Kaggle. This dataset would be used for cross-silo FL for text classification.

**Cross-Sector Dataset:** This dataset comprises three smaller consumer review datasets of different customer service-related sectors, each acting as a client: Amazon, Yelp, and Hotel [21, 26,27]. All reviews are rated 1 to 5. The datasets were obtained from Kaggle. This dataset would be used for the cross-silo FL setting for the text classification task.

**CIFAR-10 FL Dataset:** CIFAR-10 [28] is a well-known dataset for image classification. For cross-device FL, the training set is divided into 100 subsets, each acting as a client. Following [29], to simulate non-iid-ness in the dataset amongst clients, each client is allocated images from each class using a Dirichlet distribution with $\alpha = 0.9$. As for evaluating the FL model on each client, unlike other datasets where each client had its own test set, the original CIFAR-10 test set is used, with the model's per-class accuracy being multiplied by the corresponding class's ratio in the client's training set and summed up.

**Reddit Dataset:** For next-word prediction with cross-device FL, the dataset from [29], which is made up of posts of 80,000 Reddit users from November 2017, is used [30]. A corpus of the 50,000 most frequent words is used for the task, with the rest being replaced with the <unk> token. The data for each user was split into the training and testing sets in a ratio of 90:10.

### 3.2   Tasks & Model Architecture

**Image Classification**: ResNet18 model architecture [31] was used for image classification tasks, with randomly initialized weights. Stochastic gradient descent (SGD) was used as the optimizer for all experiments as most Federated Learning works currently use SGD. The metric was top-1 accuracy.

The FL model was trained for 100 rounds for cross-silo FL, with all clients participating for every round. For cross-device FL, the FL model was trained for 1000 epochs, with each round involving ten randomly selected clients. For both scenarios, local training for each client was for two epochs.

**Text Classification**: A CNN model with word embeddings for sentence classification [32] was used. FL model was trained for 20 rounds, with two epochs of local training for each client, with all clients participating every round.

**Next-Word Prediction:** For next-word prediction, a two-layer LSTM model with 200 hidden units and 10 million parameters was used, following [29]. FL model was trained for 2000 rounds with 100 randomly selected clients participating in each round. For personalization of the FL model, 8000 clients were randomly selected for personalization experiments rather than all clients.

### 3.3   Personalization Approaches

In terms of coming up with combinations of different personalization approaches, there is a need to consider the cross-compatibility of different approaches and where they augment the traditional FL process. FT is universally compatible and is the basis for the other approaches except for MoE. FB is a modification of FT, so it is universally compatible as well. MTL and KD modify FT/FB's loss function so they would be mutually exclusive in combinations. MoE would come in after local personalization is done through a combination of FT/FB and KD/MTL. As such, the combinations of personalization approaches to be explored can be found in Table 2.

**Table 2: Combinations of Approaches**

| Combination | FT | FB | KD | MTL | MoE |
|---|---|---|---|---|---|
| 1 | ✓ | | | | |
| 2 | ✓ | | ✓ | | |
| 3 | ✓ | | | ✓ | |
| 4 | | ✓ | | | |
| 5 | | ✓ | ✓ | | |
| 6 | | ✓ | | ✓ | |
| 7 | | | | | ✓ |
| 8 | ✓ | | | | ✓ |
| 9 | ✓ | | ✓ | | ✓ |
| 10 | ✓ | | | ✓ | ✓ |
| 11 | | ✓ | | | ✓ |
| 12 | | ✓ | ✓ | | ✓ |
| 13 | | ✓ | | ✓ | ✓ |

## 4 Results & Discussion

### 4.1 Local Models vs. FL Models

As mentioned in Section 1, local models trained only on individual clients' data can sometimes perform better than FL models. This is due to the FL model failing to adapt to that particular client's unique data distribution.

We expect that cross-silo FL would see a more significant discrepancy between local and FL models, with local models performing better than FL models than cross-device FL. This is due to each client in cross-silo FL having a sizeable local dataset that can be used to train decently performing local models, while each client in cross-device FL tends to have much lesser local training samples. The type of clients for cross-silo FL, which is usually large organizations, would also mean more significant statistical heterogeneity amongst clients since different organizations might use different data collection methods, software, and hardware. This contrasts with cross-device FL, where clients would tend to use the same kind of application and generate data of a more similar nature.

Our experimental results support the hypothesis above, with all cross-silo tasks having the FL model underperforming compared to local models, as seen in Figure 1. For cross-device tasks, the FL model underperformed for less than 1% of clients for image classification and next-word prediction tasks. Such underperformance for cross-silo tasks is concerning for FL, as it will disincentivize clients from joining the federation. Any client leaving or not joining the federation will have a noticeable effect since each client contributes a significant proportion of the overall dataset compared to the cross-device scenario.

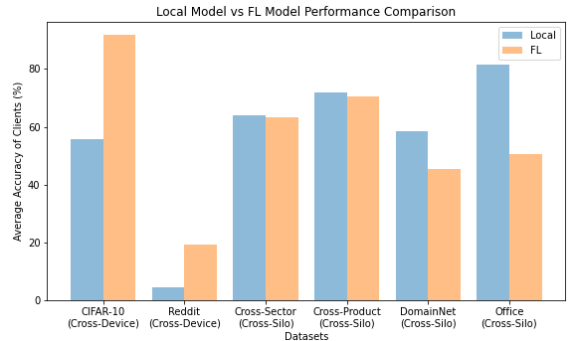

**Figure 1: Performance comparison between local and FL models**

### 4.2 FL with DP and RA

With DP and RA, performance degrades further as expected, as seen in Figure 2, which compares the average accuracy of tasks for normal FL against DP-FL and RA-FL where applicable. Results suggest that RA-FL causes more degradation of performance than DP in cross-device tasks. This could be due to non-iid data, together with median aggregation, which would take the median update that belongs to a single client, even though that client might have a skewed distribution of data.

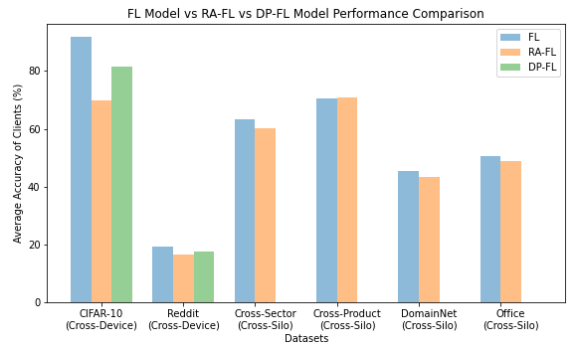

**Figure 2: Performance comparison between FL, RA-FL, and DP-FL models**

In cross-silo scenarios, degradation caused by RA-FL is not as significant. This could be due to the small number of clients involved, so there is a higher chance of a client's update being the median in each round, resulting in a greater degree of representation of each client's distribution in the eventual FL model, so average performance does not degrade as much. The results show an inherent trade-off between having enhanced privacy/integrity measures and FL models' performance.

### 4.3 Individual Personalization of FL Models

The results of the experiments described in Section 3 can be found in Table 3 and Table 4 for cross-silo and cross-device FL, respectively. For each column, the combination(s) with the highest accuracy is bolded. The subsections that follow will analyze and discuss points of interest in the results.

*4.3.1 Finetuning Recovers Performance.* As seen in Table 3 and Table 4, finetuning the FL model (FL+FT) helps the FL model's performance recover from the degradation. After finetuning, the average performance of the FL model was also better than the local models for all tasks across both cross-silo and cross-device FL. This shows that finetuning as a personalization approach works and can be universally applied. Finetuning is also simple to implement since it can make use of the same algorithm for local training during FL, and all clients would be able to implement such an approach without much difficulty. This makes finetuning something that should always be implemented along with FL in real-life applications.

*4.3.2 Mixture of Experts – Another Universal Personalization Approach.* Mixture of experts (FL+MoE) is a personalization approach that requires an additional local model to be trained for each client. With the training and evaluation algorithms and setup already implemented for FL training, the training of a local model for MoE would only require additional space and training time for clients to implement. In terms of performance gains, MoE is another personalization approach that yields improvements across both cross-silo and cross-device FL scenarios and tasks, as seen in Table 3 and Table 4. Compared to finetuning, MoE does not overwrite or customize the features learned in FL but rather complements it separately using the weighted average. So, for instances where the testing data contains features that are not commonly found in the training data, the FL model may be weighted higher, and for instances where testing data contains features found in training data, the domain expert model may be weighted higher.

*4.3.3 Performance of Knowledge Distillation and Multi-Task Learning.* Both KD and MTL implemented for this project are extensions of FT, with the loss function being the main differentiating factor between the three. For both cross-silo and cross-device FL, both approaches seem to have little or no positive impact on performance and sometimes even degrade the personalized FL model. This could be explained once again by the non-iid-ness of the datasets.

Since each client would have its own unique distribution of data and features, having approaches like KD and MTL which put further emphasis on deriving learned features from the generalized global FL model would not be useful. However, it is notable that when these approaches are combined with MoE, there seem to be better improvements in performance. This will be elaborated further on in the subsequent sections.

## 4.4 Combining Personalization Approaches

*4.4.1 Combining FT and MoE.* Earlier subsections have shown that FT and MoE are universally suitable to be implemented for all FL scenarios and tasks explored in this project. Combining these two approaches alone also gave the best performance out of all combinations tested for five out of eight cross-silo FL setups, as seen in Table 3.

Combining FT and MoE ensemble allows for the best features of both approaches to be combined, with the domain-specific features being used in cases where the domain expert model is very sure of its prediction and the expanded and finetuned feature space from the finetuned FL model being used in cases where the domain expert model is not too sure of its prediction.

*4.4.2 Combinations with FB.* For both cross-silo and cross-device FL, just like how FL+FB on its own performed worse than FL+FT, combination approaches with FB can be seen in Tables 3 and 4 to not perform well normal FT combinations beating FB combinations most of the time. Even for cases where combinations with FB performed the best, other combinations' performance is not too far off or equal to it. Hence, FB should not be used in most situations for combination approaches.

Table 3: Average accuracy (%) of clients for cross-silo FL ablation study

| Scenario | Cross-Silo | | | | | | | |
|---|---|---|---|---|---|---|---|---|
| Dataset | Cross-Sector | | Cross-Product | | DomainNet | | Office | |
| Local Model | 64.13 | | 71.76 | | 58.36 | | 81.71 | |
| Approach | FL | RA-FL | FL | RA-FL | FL | RA-FL | FL | RA-FL |
| FL | 63.44 | 60.09 | 70.43 | 70.80 | 45.40 | 43.31 | 50.48 | 48.67 |
| FT + FT | 64.58 | 64.65 | 72.61 | 72.94 | 59.45 | 59.68 | 84.31 | 82.69 |
| FL + FT + KD | 63.99 | 63.30 | 71.22 | 71.36 | 55.94 | 56.40 | 83.72 | 81.82 |
| FL + FT + MTL | 64.33 | 63.30 | 71.14 | 71.94 | 59.57 | 59.09 | 83.05 | 86.92 |
| FL + FB | 64.11 | 63.49 | 71.16 | 71.32 | 56.64 | 57.38 | 83.48 | 77.44 |
| FL + FB + KD | 63.48 | 62.89 | 70.80 | 71.16 | 56.02 | 56.14 | 83.48 | 77.91 |
| FL + FB + MTL | 63.73 | 60.74 | 70.45 | 70.82 | 56.59 | 56.80 | 83.48 | 77.91 |
| FL + MoE | 65.90 | 65.56 | 73.20 | 73.24 | 59.68 | 59.58 | 83.01 | 83.10 |
| FL + FT + MoE | **67.01** | **66.08** | **73.66** | **73.70** | 61.42 | 61.26 | **86.01** | 84.09 |
| FL + FT + KD + MoE | 65.91 | 65.12 | 73.26 | 73.20 | 61.19 | 60.66 | 85.56 | 85.68 |
| FL + FT + MTL + MoE | 66.05 | 65.60 | 73.29 | 73.51 | **61.69** | **61.43** | 85.20 | **87.15** |
| FL + FB + MoE | 66.26 | 65.64 | 73.24 | 73.30 | 61.65 | 61.01 | 85.56 | 84.54 |
| FL + FB + KD + MoE | 65.82 | 65.35 | 73.26 | 73.24 | 61.19 | 60.77 | 85.56 | 85.26 |
| FL + FB + MTL + MoE | 65.97 | 65.69 | 73.27 | 73.31 | 61.20 | 61.09 | **86.01** | 85.14 |

Table 4: Average accuracy (%) of clients for cross-device FL ablation study

| Approach | Cross-Device | | | | | |
|---|---|---|---|---|---|---|
| Dataset | CIFAR-10 | | | Reddit | | |
| Local Model | 55.65 | | | 4.27 | | |
| Approach | FL | DP-FL | RA-FL | FL | DP-FL | RA-FL |
| FL | 91.84 | 81.40 | 69.86 | 19.02 | 17.46 | 16.50 |
| FT + FT | 94.23 | 87.02 | 78.06 | 19.77 | 19.00 | 17.03 |
| FL + FT + KD | 89.48 | 80.35 | 71.46 | 20.29 | 18.69 | 17.76 |
| FL + FT + MTL | 94.16 | 87.20 | 79.00 | 20.38 | 19.18 | 17.54 |
| FL + FB | 93.77 | 87.94 | 79.37 | 20.33 | 19.06 | 17.52 |
| FL + FB + KD | 88.56 | 78.21 | 78.28 | 19.70 | 18.31 | 17.02 |
| FL + FB + MTL | 93.58 | 87.02 | 78.07 | 20.16 | 18.81 | 17.39 |
| FL + MoE | 92.70 | 84.01 | 74.14 | 19.68 | 18.22 | 16.95 |
| FL + FT + MoE | 94.38 | 87.07 | 78.30 | 20.16 | 19.32 | 17.48 |
| FL + FT + KD + MoE | 91.32 | **91.19** | 75.28 | 20.55 | 18.95 | 17.94 |
| FL + FT + MTL + MoE | **96.05** | **91.19** | 79.29 | **20.81** | **19.53** | **18.01** |
| FL + FB + MoE | 94.01 | 88.02 | **79.54** | 20.53 | 19.23 | 17.69 |
| FL + FB + KD + MoE | 93.85 | 88.02 | 78.54 | 20.08 | 18.66 | 17.25 |
| FL + FB + MTL + MoE | 93.80 | 88.02 | 73.69 | 20.50 | 19.14 | 17.63 |

*4.4.3 Combinations with KD and MTL.* Although KD and MTL on their own are not effective personalization approaches, when combined with MoE, there is a greater degree of performance improvement. This could be due to KD and MTL acting to create a personalized FL model that is more influenced by the global pool of data since both approaches introduce additional influences in the loss function based on the original global FL model. This effect would allow an even wider distribution of features to be accessible to the MoE ensemble, thus increasing performance to a greater degree than FL+FT+MoE. This effect is more pronounced in cross-device FL, with the KD/MTL combinations obtaining the best performance in five out of six setups, compared to 3 out of 8 setups for cross-silo FL.

Between KD and MTL, MTL appears to be the better personalization approach in terms of performance, with MTL outperforming KD in 39 setups, KD outperforming MTL in 12 setups, and being of equal performance in 5 setups. As such, MTL should be used in cross-device FL scenarios and tasks when possible. KD does not appear to be a viable alternative to MTL in terms of performance. For cross-silo FL scenarios and tasks, the suitability of MTL would be limited to tasks such as image classification and not text classification ones.

*4.4.4 Effect of Combination Approaches on RA-FL and DP-FL.* All combination approaches explored managed to compensate for degradation caused by the additional privacy and integrity features of RA-FL and DP-FL. Such an effect would re-incentivize clients which may not have joined or left when the setup was plain FL without personalization. These individual and combination approaches also do not incur much additional overheads in terms of resources like time and space, and the mechanism for implementing them is already mostly available through the implementation of the FL framework.

Combination approaches clearly provide a better performance gain compared to individual ones, and therefore FL framework implementations should always try to include combination approaches as part of their personalization solutions.

*4.4.5 Best Combination Approach for Cross-Silo and Cross-Device FL.* Across the eight tasks in cross-silo FL, for image classification tasks, the best combination of personalization approaches contains FT/FB, MoE, and MTL. For text classification tasks, the best combination of personalization approaches contains just FT with MoE. Across the six tasks in cross-device FL, the best combination of personalization approaches contains FT, MoE, and MTL.

## 5 Conclusions and Future Work

The success of federated learning systems is dependent on the number of clients that participate, and this is influenced by the benefits that clients get in terms of performance gains, as well as the protections such systems offer, such as privacy and integrity.

We have shown that due to the statistical heterogeneity present across clients' data and the addition of privacy and integrity protection, the performance of FL systems can suffer, sometimes to the point where non-participation is favored. Personalization of FL models, either through standalone approaches or combined ones, can reverse this performance degradation and even bring additional gains in performance, without the need for significant additional resources. Among the combinations of personalization approaches explored for both cross-silo and cross-device FL, combinations with finetuning, mixture of experts, and multi-task learning gave the best performance gains.

Future work could be in the form of further exploration of different FL tasks beyond the domains of computer vision and natural language processing, more varied personalization approaches that go beyond the typical archetypes presented here, or into the vertical FL scenarios since this project is solely focused on the horizontal FL scenario.

## Acknowledgements

This research is supported, in part, by the National Research Foundation, Singapore under its the AI Singapore Programme (AISG2-RP-2020-019); the Joint NTU-WeBank Research Centre on Fintech (NWJ-2020-008); the Nanyang Assistant Professorship (NAP); the RIE 2020 Advanced Manufacturing and Engineering Programmatic Fund (A20G8b0102), Singapore; the SDU-NTU Centre for AI Research (C-FAIR), Shandong University, China. Any opinions, findings and conclusions or recommendations expressed in this material are those of the authors and do not reflect the views of the funding agencies.